\documentclass[sigconf, nonacm]{acmart}

\usepackage{float}
\usepackage{placeins}
\usepackage{listings}
\AtBeginDocument{%
  }

\setcopyright{none}
\copyrightyear{2026}
\acmYear{2026}
\acmDOI{}
\acmISBN{}

\author{Joong Ho (Joshua) Choi}
\email{joongho.choi@bny.com}
\orcid{0009-0003-2531-5359}
\affiliation{%
  \institution{BNY}
  \city{Pittsburgh}
  \state{PA}
  \country{USA}
}

\author{Aditya Paul}
\email{aditya.paul@bny.com}
\affiliation{%
  \institution{BNY}
  \city{Pittsburgh}
  \state{PA}
  \country{USA}
}

\author{Nidhi Singh}
\email{nidhi.singh@bny.com}
\affiliation{%
  \institution{BNY}
  \city{Pittsburgh}
  \state{PA}
  \country{USA}
}
\begin{document}

\title{CARLAR Influence: A Structured Hallucination Risk in Multimodal Cheque Extraction and an Exploratory Compressed-Image Countermeasure}

\begin{abstract}
 Multimodal large language models (LLMs) are increasingly deployed for structured information extraction from financial documents, yet their failure modes under visual ambiguity remain poorly understood. This paper investigates a previously underexplored failure mode in vision-language models for financial document processing, which we term CARLAR (Courtesy Amount Recognition and Legal Amount Recognition) influence. In cheque understanding tasks, multimodal models can return agreeing monetary fields even where the numeric annotations disagree after a blank annotation is encoded as zero, producing outputs that appear internally consistent and may evade standard discrepancy checks. Through a systematic evaluation across ten visual prompting strategies applied to GPT-4.1 and Claude 4.5 Sonnet --- evaluated on 243 primary cheques (124 with two observed unequal amounts) and 598 strict mismatches from two other datasets --- with bootstrap confidence intervals and formal significance testing, we show that CARLAR rates reach 73.7\% [68.3, 79.0] under courtesy redaction ($p < 0.001$ vs.\ baseline), and that LLM confidence scores can remain high when the predicted fields agree (Expected Calibration Error up to 0.93 in CARLAR cases). To address this, we propose a lightweight compressed image few-shot remedy that encodes structured example tables as compact visual context without requiring model fine-tuning, improving Claude's joint extraction accuracy by up to 11.5 percentage points [6.2, 16.9] (paired, adjusted $p = 0.00054$) on the 243-case cohort; the strict-124 Control gain is smaller and not significant after correction. The effect on predicted-field agreement remains model-dependent, decreasing for Claude but increasing for GPT-4.1. These findings underscore the need for context-aware multimodal validation strategies in high-stakes document processing workflows.
\end{abstract}

\maketitle


\section{Introduction}

The automation of financial document processing has long been a critical application domain for AI systems. Cheque processing, in particular, represents a multi-billion-dollar annual operation across financial institutions. Cheques contain two critical amount fields: (1) the \textbf{courtesy amount}, printed digitally on the lower-right corner (numerals), and (2) the \textbf{legal amount}, written in words on the center of the cheque. These fields are legally required to match; discrepancies are red flags for fraud or processing errors.

Old neural approaches \cite{xu2020layoutlm} have helped improve accuracy in automation pipeline. While modern OCR solutions such as Azure OCR are capable of operating across diverse document types, they function primarily as transcription tools — converting visual text into machine-readable characters without inherent understanding of document semantics, field relationships, or contextual meaning. Extracting structured information from unstructured financial documents using OCR therefore still requires carefully engineered downstream pipelines: template matching, rule-based field localization, and explicit validation logic that must be maintained and updated as document formats evolve. 

The emergence of multimodal Large Language Models (such as GPT-4.1 \cite{openai_chatgpt4.1} and Claude 4.5 Sonnet \cite{claude_sonnet_4.5}), which combine vision and language understanding, has enabled directly interpreting layouts and extracting values rather than transcribing text and deferring interpretation to hand-crafted rules. These models jointly reason over visual layout and linguistic content to directly infer structured field values — understanding, for instance, that a spelled-out word amount and a numeric amount on a cheque represent the same underlying value and should be consistent. This semantic awareness allows a single model to generalize across diverse document formats without domain-specific engineering.

We study a failure signature we term \textbf{CARLAR influence}: equal predicted amounts where numeric annotations differ, including when a blank raw annotation is encoded as zero. A model may infer one amount from the other when visual evidence is ambiguous, but prediction agreement alone does not establish that mechanism. In banks, when numerically and verbally written amounts conflict, the legal amount (in words) prevails; an extraction that makes the two fields appear equal could conceal such a discrepancy.

Selective classification provides a general mechanism for deferring uncertain predictions for review~\cite{geifman2017selective}. In cheque extraction, routing low-confidence predictions or discrepancies between courtesy and legal amounts to a human operator is one possible safeguard, not a bank policy or numerical threshold established by this study. Our findings show why either signal alone can miss errors: LLMs can assign high confidence to incorrect outputs, and CARLAR cases may yield matching predictions that do not trigger discrepancy-based review.

Incorrect amount extraction carries direct financial, regulatory, and reputational risk, motivating robust validation mechanisms beyond confidence thresholding. Collectively, these risks underscore the importance of robust validation mechanisms in AI-driven payment automation systems.

To address these issues, this paper makes the following contributions:

\begin{enumerate}
    \item \textbf{Empirical Characterization of CARLAR Influence}: We systematically measure predicted-field agreement across ten visual presentation strategies on 243 cheques, including a strict sensitivity analysis on the 124 cases with two observed unequal amounts. With courtesy redacted and legal visible, agreement increases to 73.7\% [68.3, 79.0] (GPT-4.1) and 49.0\% [42.8, 55.6] (Claude 4.5).
    
    \item \textbf{Failure of Confidence-Based Defenses}: Using formal calibration metrics (ECE, Brier score, risk-coverage curves, and selective accuracy), we demonstrate that LLM confidence scores are poor proxies for extraction fidelity, with the model maintaining high confidence on CARLAR-affected outputs. This challenges the assumption that confidence thresholding alone suffices for safe deployment.
    
    \item \textbf{Compressed Image $n$-Shot Remedy}: We evaluate structured example tables as compact visual context; the strongest paired joint-accuracy gain is 11.5 percentage points [6.2, 16.9] for Claude 4.5 with Control examples (BH-adjusted $p = 0.00054$), without model fine-tuning. GPT-4.1's Prediction-Guided and Comprehensive changes are not significant in the 243-case six-test family, while Claude's Control gain is not significant on the smaller strict-124 subset after separate adjustment. The example sets' effect on measured CARLAR is model-dependent: the verified example sets reduce predicted-field agreement for Claude 4.5 but increase it for GPT-4.1, both in the primary cohort and in the strict two-field subset. These observations do not identify the causal direction of any individual prediction.
    
    \item \textbf{Multimodal Error Analysis}: We show that CARLAR influence varies with visual context and input representation choices. Redacting courtesy while preserving legal produces substantially more predicted-field agreement than the reverse intervention; Relative Deviation Asymmetry (RDA) characterizes numeric drift toward the other annotated amount that exact-match metrics miss. Per-sample persistence profiling across all ten experiments shows that 70\% of samples are condition-sensitive rather than intrinsically vulnerable.
\end{enumerate}

\section{Related Work}

\subsection{LLMs for Document Understanding}

Document-VQA benchmarks such as DocVQA~\cite{mathew2021docvqa} established evaluation tasks for document images before recent multimodal LLM extraction. Complementary foundations include few-shot in-context learning in language models~\cite{brown2020gpt3}, text-to-text transfer learning~\cite{raffel2020t5}, and learned prompts for CLIP-like vision-language models~\cite{zhou2022coop}. These studies do not directly evaluate cheque extraction or the failure modes that arise when related visual fields disagree.

The term ``hallucination'' in LLM literature refers to generating factually incorrect or unsupported outputs~\cite{huang2025hallucination}. Object hallucination is a documented failure in large vision-language models: generated descriptions can name objects absent from an image~\cite{li2023visionhallucination}. In financial LLM evaluation, objective bias can arise when training or evaluation rewards confident completion rather than calibrated uncertainty and abstention~\cite{kong2026evaluating}. A similar pressure could contribute to unsupported agreement between cheque fields, but neither that finance-evaluation study nor the CARLAR metric establishes such a mechanism in visual question answering. Predicted-field agreement is consistent with a possible relational error but can also reflect other errors or common defaults. Our single-pass experiments show condition sensitivity rather than determinism: we did not repeat identical image--prompt inputs to measure within-condition variability. 

Task-aware image captions have been used to provide relevant visual context for language-model VQA~\cite{hu2023promptcap}, and prompting techniques have been studied for zero- and few-shot VQA~\cite{awal_vqa_prompting_2023}. These works motivate studying how context affects extraction, but do not evaluate our compressed example-table format or cheque-specific field agreement. 

Confidence-based review has limitations. Verbalized uncertainty in language and vision-language models can be overconfident~\cite{groot2024overconfidence}, although language models can sometimes assess whether their responses are correct~\cite{kadavath2022confidence}. TruthfulQA documents false answers that reflect common human misconceptions~\cite{lin2022truthfulqa}; calibration of image classifiers is a distinct research setting~\cite{minderer2021confidence}, not direct evidence about LLM self-reported confidence. LLM-as-judge assessments can also diverge from human evaluations~\cite{thakur2025judging}.

To our knowledge, no prior work has systematically studied hallucination modes in LLM-based cheque processing specifically, nor explored how additional image context of few-shot examples can contribute. In addition to covering these gaps, our paper also proposed new defense mechanisms tailored to financial workflows beyond the conventional practice of having additional humans in the loop or LLMs as judge. 

\section{Proposed Intervention: A Compressed Image Few-Shot Remedy}

A natural question is whether CARLAR influence could be mitigated 
using text-only few-shot prompting. However, CARLAR arises specifically 
under visual ambiguity — illegible handwriting, occlusion, or poor 
scans — conditions that cannot be faithfully represented in text without 
resolving the ambiguity or introducing artificial placeholders. Our 
compressed image few-shot approach instead preserves the original visual 
signal while providing structured contextual examples, ensuring mitigation remains grounded in the multimodal setting where CARLAR occurs.

To mitigate CARLAR influence without model fine-tuning, we propose a lightweight compressed image few-shot remedy that incorporates structured examples as compact visual context.

\textbf{Step 1 — Example Selection}: We first identify a set of $n$ representative examples to guide the model. These examples are chosen based on their relevance to the target task and the specific failure modes of interest.

\textbf{Step 2 — Example Compression}: As a design motivation, we explore whether the selected examples can be supplied as one visual table rather than as $n$ separate image inputs. We organize them into a structured table, where each row corresponds to an example and columns capture relevant fields (e.g., courtesy amount, legal amount, and ground truth). We then convert this table into a single image by capturing a screenshot of the first page, which contains all $n$ examples (Figure~\ref{fig:nshot_Table}). This representation provides compact visual context without establishing a resource saving.

\textbf{Step 3 — Multimodal Integration}: During inference, we prepend the table screenshot as an additional input to the multimodal LLM, followed by the query document. This lets the model condition on multiple examples from one table image. We did not benchmark token usage, billed API charges, or latency against separate example images; potential efficiency is a motivation, not a measured result.

\section{Design}
\subsection{Dataset}

We evaluate our approach using three publicly available cheque datasets.

\textbf{Primary dataset.} The handwritten cheque VQA dataset introduced by aniketVerma07~\cite{verma2024handwrittenchequevqa} contains 3,312 question--answer pairs spanning cheque formats from the U.S., Singapore, and India. The original amount-query filter selected 252 rows from its 3{,}000-row train split; nine corresponded to non-cheque documents and are excluded from all reported cheque analyses. Of the 243 remaining cheques, 124 have two nonblank, unequal amount annotations; the other 119 have a blank raw annotation (72 courtesy, 47 legal), represented as zero for evaluation. An audit found no nonblank annotations that failed the amount parser among these 119. A blank annotation does not establish that the corresponding writing is physically absent from every image. We report the 243-case unequal-or-blank-annotation cohort for detailed experiments and the strict-124 subset separately for comparability with the other datasets. This is a selected subset of the dataset's train split, not its held-out test split.

\textbf{Cross-dataset validation.} To assess whether CARLAR generalizes beyond a single dataset, we additionally evaluate the Experiment~1 (Full Image) baseline on two independent cheque corpora:
\begin{itemize}
    \item \textbf{arunchincheit}~\cite{arunchincheit2024cheque}: 2{,}500 images of Indian bank cheques (ICICI, AXIS, SBI, HSBC), yielding 200 mismatched samples after filtering.
    \item \textbf{shivalikasingh}~\cite{shivalikasingh2024cheque}: 2{,}800 images of Indian bank cheques, yielding 398 mismatched samples after filtering.
\end{itemize}
Together these three datasets provide 841 evaluated cheques: 243 primary unequal-or-blank-annotation cases and 598 external cases with two observed unequal amounts. The strict subset contains 722 two-field mismatches across all three datasets (124 + 200 + 398). The detailed visual experiments and compressed image few-shot comparisons use the primary 243 cheques; the two additional Indian cheque corpora are evaluated only for the full-image baseline. Cross-dataset interpretation uses the strict primary subset, rather than treating blank raw annotations as equivalent to two observed conflicting amounts.

Although these subsets are small and domain-specific by design, predicted-field agreement is particularly concerning when two amounts are known to differ. The primary cohort also includes cheques with a blank raw annotation; these evaluate responses to an unlabeled field under the zero convention, not substitution between two observed annotated amounts. These cases reflect ambiguity patterns encountered in proprietary banking workflows where CARLAR was originally identified. Due to confidentiality constraints, we cannot release internal datasets, but the public subsets serve as representative, reproducible testbeds rather than claims of broad coverage.

\subsection{Method}

We design ten experiments to systematically explore the conditions under which CARLAR influence emerges. Each experiment varies the visual presentation of the cheque and the extraction strategy.

1) \textbf{Full Image}: The unmodified 512$\times$256 px cheque image. This serves as baseline. 

2) \textbf{Annotated Both}: Full image with red bounding box around legal amount (center, word-based) and green bounding box around courtesy amount (lower-right, numeric). Provides unambiguous spatial guidance via color.

3) \textbf{Annotated Separate}: Two separate calls—one with red box on legal region only (prompt: ``extract legal amount''), one with red box on courtesy region only (prompt: ``extract courtesy amount''). Tests whether isolating each extraction reduces CARLAR influence.

4) \textbf{Cropped Separate}: Two separate calls—one with tight crop of legal region, one with tight crop of courtesy region. Tests whether removing visual context (other fields, cheque borders) helps focus attention. Cropping coordinates: Legal (0, 80, 430, 145), Courtesy (350, 95, 510, 150).

5) \textbf{Cropped Courtesy + Full Legal}: Asymmetric—courtesy amount extracted from cropped image, legal amount from full image. Tests whether different fields benefit differently from cropping.

6) \textbf{Cropped Legal + Full Courtesy}: Inverse asymmetry.

7) \textbf{Redacted Separate}: Two separate calls—one receives full image with legal region blacked out (prompt: ``ignore the black box''), one receives full image with courtesy region blacked out. Tests whether providing context while redacting the relevant field helps.

8) \textbf{Redacted Courtesy + Full Legal}: Asymmetric redaction—courtesy extracted from redacted image, legal from full image.

9) \textbf{Redacted Legal + Full Courtesy}: Asymmetric redaction—inverse.

10) \textbf{Gridlines}: Full image with 16$\times$8 numbered grid overlay (128 cells, $\approx 32 \times 32$ px each). Prompt explicitly specifies which grid cells contain each amount. Tests textual spatial guidance vs. visual/color-based guidance.

For model selection, we chose to evaluate two state-of-the-art multimodal LLMs: GPT-4.1 and Claude 4.5 Sonnet. Both models support multimodal in-context learning and have demonstrated strong document understanding capabilities in preliminary tests. We were not able to run Gemini for ablations due to our firm's closer commercial relationships with Anthropic and OpenAI.

For each experiment, we craft scenario-specific prompts to extract courtesy and legal amounts into JSON; confidence is elicited in separate confidence experiments. We evaluate field accuracy and predicted-field agreement on the 243 selected cheque images. Agreement on cases with two known unequal amounts is a stronger signal of cross-field substitution than agreement when one raw annotation is blank; neither measure alone proves that one amount was copied from the other. 

Let $C_i^{\ast}$ and $L_i^{\ast}$ denote the ground-truth courtesy and legal amount fields for row $i$, respectively, and let $\hat{C}_i$ and $\hat{L}_i$ denote the corresponding model predictions.

\[
\mathrm{CARLAR\%}
= \frac{1}{|\mathcal{S}|}
\sum_{i \in \mathcal{S}}
\mathbf{1}\!\left[\, \hat{C}_i = \hat{L}_i \,\right]
\]

where $\mathcal{S}$ contains 243 primary cheques whose annotated values differ after blank raw annotations are encoded as zero. The 124-case strict subset requires both original annotations to be present and unequal; external datasets use that same criterion. Each cohort-specific rate uses its own stated cohort size as the denominator.

This formulation measures unexpected agreement between the two predicted fields, complementing ordinary prediction-error metrics. Standard metrics such as partial string match, edit distance, or numeric deviation are unsuitable for this purpose because they quantify proximity to ground truth, not dependency between fields. To illustrate why this distinction matters, consider a case where the ground-truth legal amount is 20 and the courtesy amount is 17, but the model predicts the courtesy amount as 20 --- for instance, due to illegible handwriting causing the model to default to the legal field. A numeric deviation metric would treat this prediction as plausible, since 20 is close to 17, and would fail to distinguish it from ordinary extraction noise. CARLAR, by contrast, flags agreement between predictions ($\hat{C}_i = \hat{L}_i$) despite unequal numeric ground truth, including blank annotations encoded as zero ($C_i^{\ast} \neq L_i^{\ast}$ under the zero convention). It is a diagnostic indicator of possible cross-field contamination; the strict-subset analysis isolates cases where both ground-truth amounts are actually observed.

We additionally define CARLAR-NZ (Non-Zero CARLAR) as the percentage of the same eligible cohort for which the model predicts equal, nonzero values for both fields:

\begin{equation}
    \mathrm{CARLAR\text{-}NZ\%} = \frac{1}{|\mathcal{S}|} \sum_{i \in \mathcal{S}}
    \mathbf{1}\!\left[\hat{C}_i = \hat{L}_i \neq 0\right].
\end{equation}

\noindent Unlike a rate conditional on two nonzero predictions, this measure has the same denominator as CARLAR\%. It removes agreement caused by two zero predictions, which may reflect abstention or a blank annotation rather than a nonzero assertion. Nonzero agreement is stronger evidence of a committed prediction, but does not on its own establish the direction or cause of substitution.

To move beyond binary detection of cross-field agreement, we introduce a complementary fine-grained metric.

\textbf{RDA (Relative Deviation Asymmetry).}
CARLAR captures only \textit{exact} agreement between predicted fields; partial drift --- where the model nudges one prediction toward the other field's value without fully copying it --- goes undetected. To capture this gradient, we define:
\[
\mathrm{RDA}_i = \frac{\hat{C}_i - C_i^{\ast}}{L_i^{\ast} - C_i^{\ast}}
\]
computed over non-CARLAR samples with non-zero predictions and $L_i^{\ast} \neq C_i^{\ast}$. An RDA of 0 means no drift (courtesy prediction matches its ground truth); an RDA of 1 means the courtesy prediction has drifted fully to the legal ground truth value. A positive RDA indicates that a courtesy prediction lies toward the legal ground-truth amount on the numeric scale; we use a Wilcoxon signed-rank test to assess departure from zero, without interpreting it as direct evidence of copying.

\subsection{Statistical Methods}

Because our per-dataset evaluation sets are modest ($n = 243$ for the primary cohort, $n = 124$ for its strict subset, and $n = 200$ and $n = 398$ for cross-dataset validation), we supplement the visual-experiment CARLAR rates with 95\% bootstrap confidence intervals (10{,}000 resamples, percentile method) and use formal hypothesis testing to avoid over-interpreting small differences.

\textbf{Pairwise comparisons.}
For paired binary outcomes (correct/incorrect, CARLAR/non-CARLAR)
across two experimental conditions on the same 243 cheques, we use
McNemar's test (exact binomial variant when the number of discordant
pairs $< 25$).
For GPT-4.1 versus Claude 4.5 within each visual condition, we align the same 243 cheque IDs and apply the exact two-sided McNemar/binomial test to the discordant CARLAR outcomes. This uses image-level pairing rather than treating the models' predictions as independent samples.
Confidence-score distributions (LLM-as-Judge vs.\ Azure OCR) are
compared with the Wilcoxon signed-rank test.
The 68 visual-condition, model-comparison, and confidence-distribution tests form one Benjamini--Hochberg-adjusted family. For six model-by-example-set comparisons of joint accuracy against prompt-v1, we align the same 243 cheque IDs, bootstrap paired correctness differences (10{,}000 resamples), use exact two-sided McNemar/binomial tests on discordant pairs, and adjust those six $p$-values as a separate Benjamini--Hochberg family. We repeat these six comparisons on the 124 strict two-field mismatches as a separate exploratory sensitivity family, with its own six-test adjustment. Raw and adjusted test results are retained in the analysis output. RDA Wilcoxon tests are exploratory and unadjusted, not multiplicity-controlled evidence.

\textbf{Calibration metrics.}
To formally evaluate whether model-reported confidence tracks
extraction correctness, we compute the \textit{Expected Calibration
Error} (ECE)~\cite{naeini2015ece, guo2017calibration} with 10 equal-width bins for the full 243-case cohort and five bins for the smaller run-specific CARLAR subsets, and the \textit{Brier score}~\cite{brier1950verification}
$\mathrm{BS} = \frac{1}{N}\sum_{i}(c_i - y_i)^{2}$, where $c_i$ is
the reported confidence (scaled to $[0,1]$) and $y_i \in \{0,1\}$ is
the binary correctness indicator. The CARLAR-only cohorts in Table~\ref{tab:calibration} contain 22--55 observations per model and confidence method; their five-bin ECE values are descriptive and cannot be interpreted as directly comparable changes from ten-bin overall ECE.
We further report \textit{risk-coverage curves} and
\textit{selective accuracy} at key confidence thresholds to
quantify how effectively confidence can filter unreliable
extractions.

For evaluation of the compressed image few-shot remedy, we construct three example table variants, each drawn from the broader public dataset (3{,}000-row train split of the Hugging Face cheque VQA corpus). The example image IDs were checked against the 243 evaluation image IDs and had no intersection; each API call receives a single static table image prepended to the query cheque.

\begin{enumerate}
    \item \textbf{Control Example Set} ($n{=}10$): Contains ground-truth amounts, model predictions, and human reasoning for 10 cheques --- 5 with matching courtesy/legal amounts and 5 with mismatched amounts --- providing diverse extraction patterns without targeting specific failure modes (Figure~\ref{fig:nshot_Table}).
    \item \textbf{Prediction-Guided Example Set (Type 1)} ($n{=}7$): Contains cheque images, model predictions, confidence scores (1--5 scale), and reasoning for seven examples with varying visual clarity. It omits ground-truth amounts, testing whether prediction-and-feedback context alone changes extraction. This is distinct from the separate Error-Similar confidence-elicitation prompts discussed later.
    \item \textbf{Comprehensive Example Set} ($n{=}5$): Contains ground-truth amounts, model predictions, confidence scores, and reasoning for 5 cheques with mismatched amounts, providing the richest per-example context.
\end{enumerate}

\noindent Because all examples must fit in a single legible table image (Step~2), the number of examples $n$ varies inversely with per-example information density: the Comprehensive variant (5 columns per row) accommodates $n{=}5$, while the Control variant (4 columns, shorter reasoning) fits $n{=}10$. This reflects the practical breadth-vs-depth tradeoff inherent to the compressed-image approach --- richer per-example context necessarily reduces the number of examples that remain legible in one screenshot.

\section{Results}

We evaluate the primary experiments on 243 cheque records from the selected train split: 124 with two nonblank unequal annotations and 119 with one blank raw annotation (72 courtesy, 47 legal), encoded as zero. CARLAR\% measures unexpected predicted-field agreement under the evaluation convention in Section~4.2.

\subsection{Understanding cases of CARLAR}
Prior to applying the three-step remedy to the 243 cheques, we manually examined selected CARLAR cases from Experiment 1 (Full Image, single call). The examples below illustrate recurring visual ambiguities in the courtesy field that can coincide with agreement between predictions. These examples do not establish a single mechanism for every agreement, particularly where a raw ground-truth annotation is blank.

\subsubsection{Whitespace CARLAR}
\begin{figure}[H]
    \centering
    \includegraphics[width=0.75\linewidth]{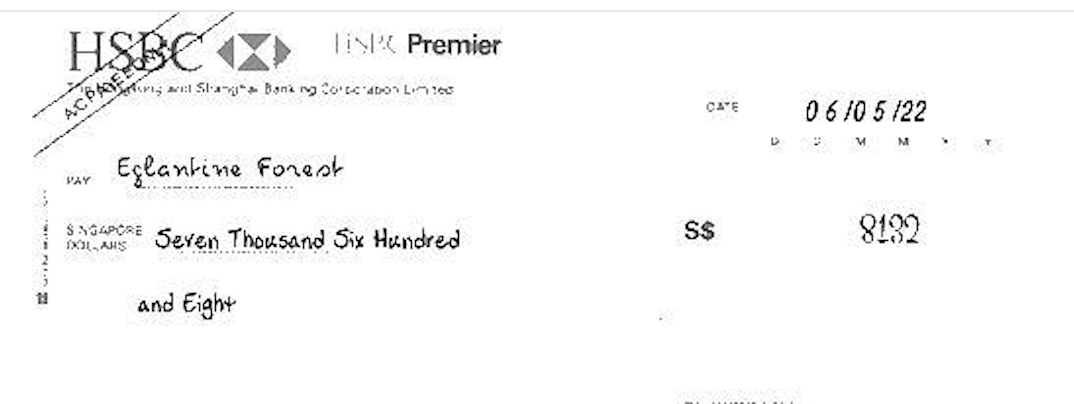}
    \caption{Whitespace in a CARLAR example}
    \label{fig:CARLAR3}
\end{figure}

Layout irregularities or spacing artifacts -- for example, unusual whitespace between the currency symbol and the numeric value (e.g., between \$ and the digits) -- appears in a CARLAR case in Fig~\ref{fig:CARLAR3}.

\subsubsection{European-style 7 CARLAR}
\begin{figure}[H]
    \centering
    \includegraphics[width=0.75\linewidth]{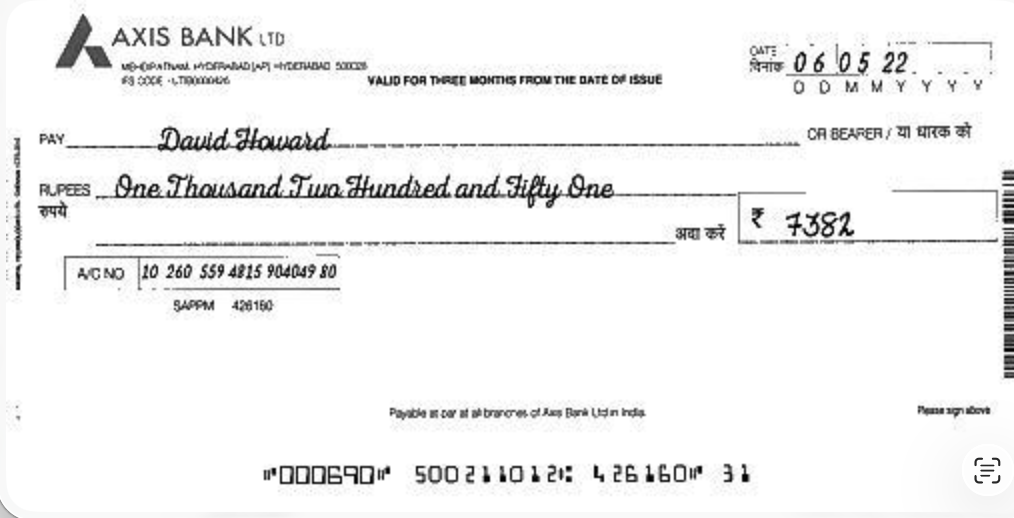}
    \caption{Handwritten "7" deviating from typical digit representation}
    \label{fig:CARLAR1}
\end{figure}

Cursive or stylized handwriting is another source of visual ambiguity in the examined cases, including European-style "7"s (Fig~\ref{fig:CARLAR1}).

\subsubsection{Stick-like 7 CARLAR}
\begin{figure}[H]
    \centering
    \includegraphics[width=0.75\linewidth]{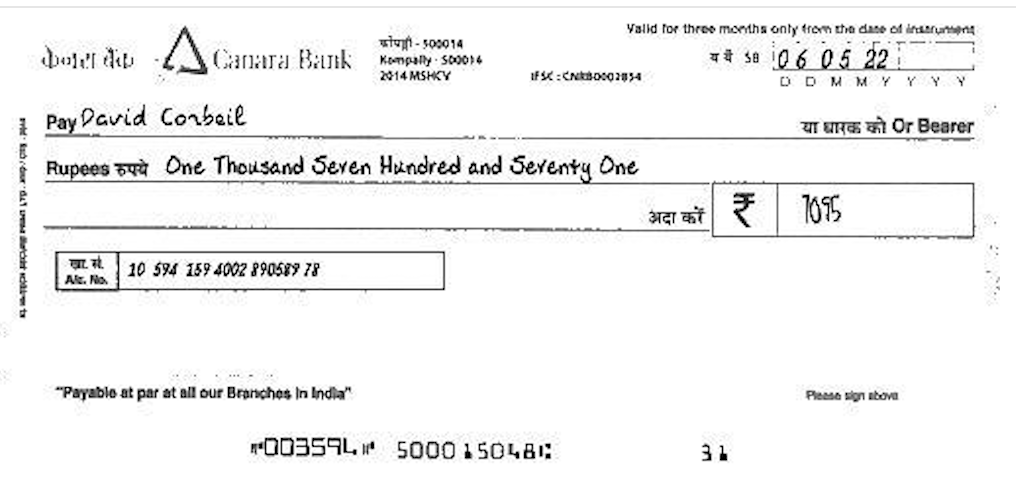}
    \caption{Digit morphology}
    \label{fig:CARLAR2}
\end{figure}

Another observed ambiguity is digit morphology reduced to a simple vertical stroke or "stick-like" figure, which is difficult to distinguish from noise or background marks (Fig~\ref{fig:CARLAR2}).

\subsubsection{Thickness/Absence Illegibility CARLAR}
\begin{figure}[H]
    \centering
    \includegraphics[width=0.75\linewidth]{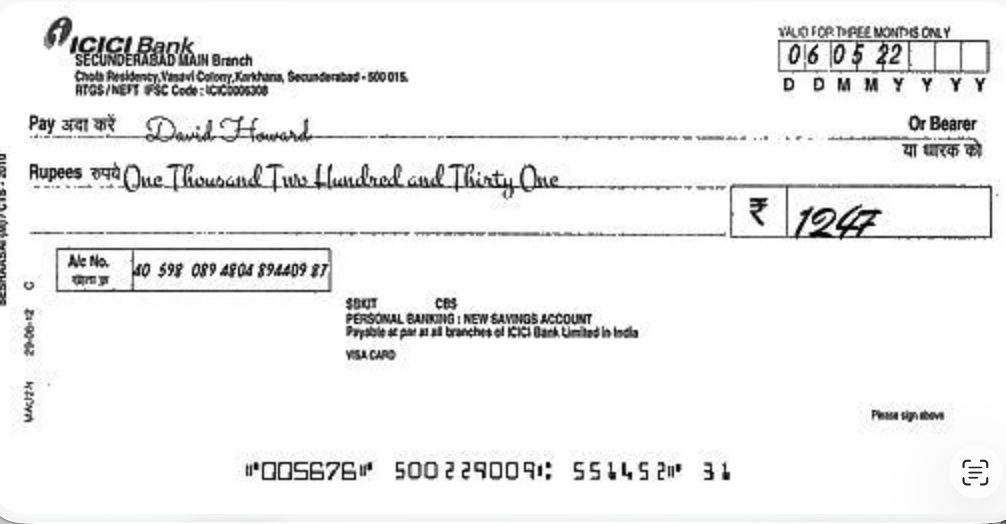}
    \caption{Unreadable courtesy amount from thick handwriting.}
    \label{fig:CARLAR4}
\end{figure}

Finally, the courtesy amount may be missing or effectively unreadable due to excessive thickness, overlap, or poor scan quality, causing the numeric field to be partially or completely lost during visual interpretation (Fig \ref{fig:CARLAR4}).

% \begin{figure}[H]
%     \centering
%     \includegraphics[width=0.75\linewidth]{CARLAR5.png}
%     \caption{Missing courtesy amount}
%     \label{fig:CARLAR5}
% \end{figure}

\subsection{Mitigating CARLAR across different methodologies}

\begin{table*}[t]
\centering
\small
\begin{tabular}{|l|c|c|c|c|}
\hline
\textbf{Experiment} & \textbf{Claude 4.5} & \textbf{95\% CI} & \textbf{GPT-4.1} & \textbf{95\% CI} \\
\hline
Exp 1 (Baseline: Full image) & 5.8\% & [2.9, 9.1] & 5.3\% & [2.9, 8.2] \\
\hline
Exp 2 (Annotated Both) & 9.9\% & [6.2, 13.6] & 14.4\% & [10.3, 18.9] \\
\hline
Exp 3 (Annotated Separate) & 4.9\% & [2.5, 7.8] & 14.8\% & [10.3, 19.3] \\
\hline
Exp 4 (Cropped Separate) & 5.3\% & [2.9, 8.2] & 2.9\% & [0.8, 5.3] \\
\hline
Exp 5 (Cropped Courtesy + Full Legal) & 1.6\% & [0.4, 3.3] & 0.4\% & [0.0, 1.2] \\
\hline
Exp 6 (Cropped Legal + Full Courtesy) & 15.2\% & [10.7, 19.8] & 16.0\% & [11.5, 20.6] \\
\hline
Exp 7 (Redacted Separate) & 13.6\% & [9.5, 18.1] & 19.8\% & [14.8, 24.7] \\
\hline
Exp 8 (Redacted Courtesy + Full Legal) & \textbf{49.0\%} & [42.8, 55.6] & \textbf{73.7\%} & [68.3, 79.0] \\
\hline
Exp 9 (Redacted Legal + Full Courtesy) & 5.8\% & [3.3, 9.1] & 6.2\% & [3.3, 9.5] \\
\hline
Exp 10 (Gridlines) & 24.3\% & [18.9, 29.6] & 39.1\% & [32.9, 45.3] \\
\hline
\multicolumn{5}{|c|}{\textit{Primary Strict Subset ($n{=}124$; Two Observed Unequal Amounts)}} \\
\hline
Exp 1 (Full image, strict) & 4.0\% & [0.8, 8.1] & 9.7\% & [4.8, 15.3] \\
\hline
Exp 8 (Redacted courtesy, strict) & 52.4\% & [43.5, 61.3] & 71.0\% & [62.9, 79.0] \\
\hline
\multicolumn{5}{|c|}{\textit{Cross-Dataset Validation (Exp 1: Full Image)}} \\
\hline
arunchincheit ($n{=}200$) & 4.0\% & [1.5, 7.0] & 6.0\% & [3.0, 9.5] \\
\hline
shivalikasingh ($n{=}398$) & 4.8\% & [2.8, 7.0] & 5.0\% & [3.0, 7.3] \\
\hline
\end{tabular}
\caption{Predicted-field agreement on 243 primary cheques (124 strict two-field mismatches and 119 blank-annotation cases), with 95\% bootstrap confidence intervals (10{,}000 resamples). The strict rows isolate two observed unequal values; external rows use separate Indian cheque datasets. The displayed primary rates use the blank-annotation-as-zero convention, and paired-condition comparisons use McNemar's test.}
\label{tab:carlar_rates}
\end{table*}

\textbf{Cross-dataset generalizability.}
The strict-primary baseline (9.7\% GPT, 4.0\% Claude; $n=124$) uses the same two-observed-unequal-amount criterion as the two independent Indian cheque datasets ($n=200$ and $n=398$). The latter have CARLAR rates of 5.0--6.0\% for GPT-4.1 and 4.0--4.8\% for Claude 4.5. These baseline results show that predicted-field agreement is not restricted to the primary dataset, although the wider primary cohort (5.3\% GPT, 5.8\% Claude; $n=243$) includes blank raw annotations and is not directly comparable. We have not evaluated the intervention or fine-grained ablations on the external cohorts.

From Table~\ref{tab:carlar_rates}, we note several findings within the primary dataset's detailed ablations. CARLAR rates depend strongly on which field is obscured: redacting courtesy while showing legal yields 49.0\% for Claude and 73.7\% for GPT (roughly 8--14$\times$ their inclusive baselines), whereas the reverse yields 5.8\% and 6.2\% (Exps~8--9). McNemar's test finds Exp~8 significantly different from baseline for both models ($p < 0.001$, Benjamini--Hochberg corrected), while Exp~9 is not (adjusted $p > 0.05$). A paired exact McNemar test also finds a between-model difference in Exp~8 (94 GPT-only versus 34 Claude-only CARLAR cases; Benjamini--Hochberg-adjusted $p = 2.38\times 10^{-7}$ within the 68-test family). On the strict subset, Exp~8 yields 52.4\% and 71.0\% agreement, respectively, so the redaction effect is not solely due to blank-annotation cases. These rates do not by themselves establish which ground-truth amount was copied in each case.

\textbf{Ruling out alternative explanations.}
A natural concern is whether predicted field agreement could arise from mechanisms other than cross-field inference — for example, correlated OCR-level errors, normalization to a shared default, or independent reliance on shared contextual cues (layout, borders, background). Experiments~4--9 vary the \emph{visibility} and visual presentation of the amount fields, allowing us to compare how predicted-field agreement changes across cropping and redaction conditions without treating those interventions as perfectly controlled counterfactuals.

\textit{Correlated OCR errors.} If both fields suffered correlated pixel-level misreads, the CARLAR rate should remain roughly constant regardless of whether the other field is visible in the input. Yet Experiment~5 (courtesy extracted from a tight crop with \emph{no} legal field visible) yields only 0.4--1.6\% CARLAR, whereas the full-image baseline (Exp~1) shows 5.3--5.8\%. More strikingly, Experiment~8 (courtesy redacted but legal fully visible) drives CARLAR to 49.0--73.7\%. The large changes in agreement rates across these conditions are difficult to attribute to independent pixel-level reading errors alone, although the conditions also change how the image is presented.

\textit{Normalization to a shared default.} If both fields were independently normalized to a common value (e.g., rounding or format coercion), agreement rates should be roughly symmetric with respect to which field is visible. Instead, Exp~8 (legal visible, courtesy redacted) yields 49.0--73.7\%, while Exp~9 (courtesy visible, legal redacted) yields only 5.8--6.2\%. This asymmetry is not what field-independent normalization alone would predict, but the agreement metric does not identify which ground-truth value, if either, was copied.

\textit{Shared contextual cues.} To isolate whether shared visual context (cheque borders, layout, background patterns) independently anchors both predictions to the same value, we compare cropping (which removes all surrounding context) against redaction (which preserves it). Experiment~5 (courtesy cropped, no surrounding context) and Experiment~8 (courtesy redacted, full surrounding context preserved) both deny the model direct visual access to the courtesy amount. Yet CARLAR jumps from 0.4--1.6\% to 49.0--73.7\%. Experiment~8 leaves the legal field visible in a full image, while Experiment~5 removes it from the courtesy crop; the difference is consistent with cross-field influence but is also confounded by cropping versus redaction.

We acknowledge that these manipulations approximate rather than perfectly implement pixel-level counterfactual controls: cropping changes image dimensions, and redaction introduces a salient black region that may itself influence model behavior (as discussed in our cropping-vs-redaction analysis below). Nevertheless, the converging evidence across three distinct manipulation types supports the interpretation that access to the other field can influence predicted-field agreement, while the redaction and cropping conditions do not isolate every possible confound.

Secondly, bounding box annotations do not seem to be effective in mitigating CARLAR. Experiment 2 (legal annotated with a red bounding box, courtesy with a green bounding box) shows higher CARLAR agreement (9.9\% for Claude and 14.4\% for GPT) than baseline (5.8\% and 5.3\%, respectively). Even when the two fields are distinguished by colour, explicitly highlighting both may prime the model to assume they should match.

Thirdly, cropping out fields seems to be an unstable way of mitigating CARLAR. Experiments 4 (separate crops) and 5 (courtesy cropped, legal full image) show lower CARLAR rates (0.4--5.3\%), but Experiment 6 (legal cropped, courtesy full image) shows \textbf{15.2\% (Claude) and 16.0\% (GPT)}. The difference suggests that the field chosen for cropping and the context retained for each extraction matter; it does not establish which field supplied the agreed value.

Fourthly, we observe that redaction triggers CARLAR influence more than cropping. This may be due to redaction (black box) preserving visual context (full cheque layout, borders, fields); the model knows a field \textit{exists} but is obscured, triggering inference. On the other hand, cropping removes context entirely, so the model may not even be aware that a field is missing; it simply processes the cropped region. Thus, redaction seems to cue the model to infer missing information, whereas cropping suggests the presented region constitutes the entire context.
    
Moreover, gridlines do not mitigate CARLAR in this setting. Experiment 10 (gridlines with textual cell references) yields 24.3\% CARLAR for Claude and 39.1\% for GPT, higher than both the full-image baseline (5.8\% and 5.3\%) and the annotated-both condition (9.9\% and 14.4\%), though lower than the courtesy-redacted condition (49.0\% and 73.7\%). These rates do not support a claim that gridline-based textual spatial guidance is more robust than color-coded bounding boxes.

Lastly, we explore the ordering and length of amount descriptions in the extraction prompt. Among four prompt variants on the same 243 cheques, Shorter Courtesy has the lowest CARLAR rate for GPT-4.1 (2.1\%, 5 cases) and is tied for the lowest rate for Claude (5.8\%, 14 cases). Legal First has the highest rate for GPT-4.1 (6.2\%, 15 cases), but not for Claude, whose rates range from 5.8\% to 7.0\%. These descriptive differences suggest that prompt wording can matter, but do not establish a common sequence effect across models.

\subsection{Fine-Grained Cross-Field Influence Analysis}

The binary CARLAR metric detects exact predicted-field agreement. We now apply RDA and persistence analysis to characterize partial drift and sensitivity to experimental conditions.

\begin{figure}[t]
    \centering
    \includegraphics[width=\columnwidth]{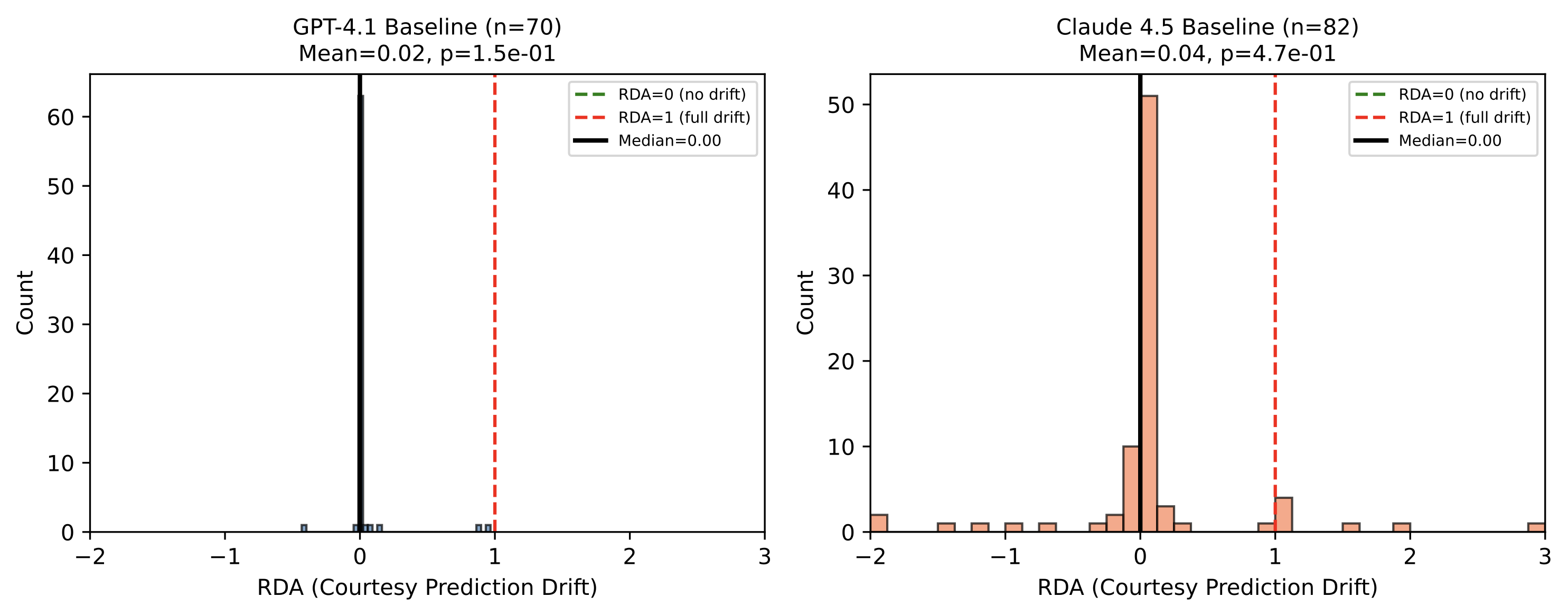}
    \caption{Distribution of Relative Deviation Asymmetry (RDA) for courtesy predictions in Experiment~1 (Full Image baseline). RDA\,=\,0 indicates no drift from courtesy ground truth; RDA\,=\,1 indicates full drift to the legal ground truth value. Black vertical line marks the median. Sample sizes differ between models because the RDA filter excludes CARLAR cases (exact agreement) and zero predictions, which occur at different rates for each model.}
    \label{fig:rda_histogram}
\end{figure}

\textbf{Partial drift (RDA).}
CARLAR captures only exact agreement between fields; a courtesy prediction can also lie partway toward the annotated legal value without matching the predicted legal amount. RDA measures this gradient: a value of 0 means no drift, while 1 means the courtesy prediction has shifted entirely to the legal ground truth (Figure~\ref{fig:rda_histogram}).

At baseline (Exp~1), neither model shows evidence of systematic drift in this exploratory analysis: GPT-4.1 has mean RDA 0.023 (unadjusted Wilcoxon $p = 0.155$), and Claude 4.5 has mean 0.040 (unadjusted $p = 0.469$). Under Experiment~7 (both fields redacted separately), the median RDA reaches 1.0 for both models (unadjusted $p < 0.001$), even among non-CARLAR predictions. These RDA tests were not included in the pairwise Benjamini--Hochberg family. RDA uses the blank-annotation-as-zero convention of the primary analysis; its values describe numeric proximity, not proof of copying from the other field.

\textbf{CARLAR persistence.}
A natural question is whether CARLAR is a property of specific cheque images or of the experimental condition applied to them. Across the same 243 cheques in all ten experiments, 73.3\% of GPT cases and 68.7\% of Claude cases show CARLAR in only one to three conditions. No cheque exhibits CARLAR in eight or more conditions for either model. This pattern suggests that agreement often depends on the image--condition interaction rather than persisting uniformly across inputs.

\begin{figure}[t]
    \centering
    \includegraphics[width=\columnwidth]{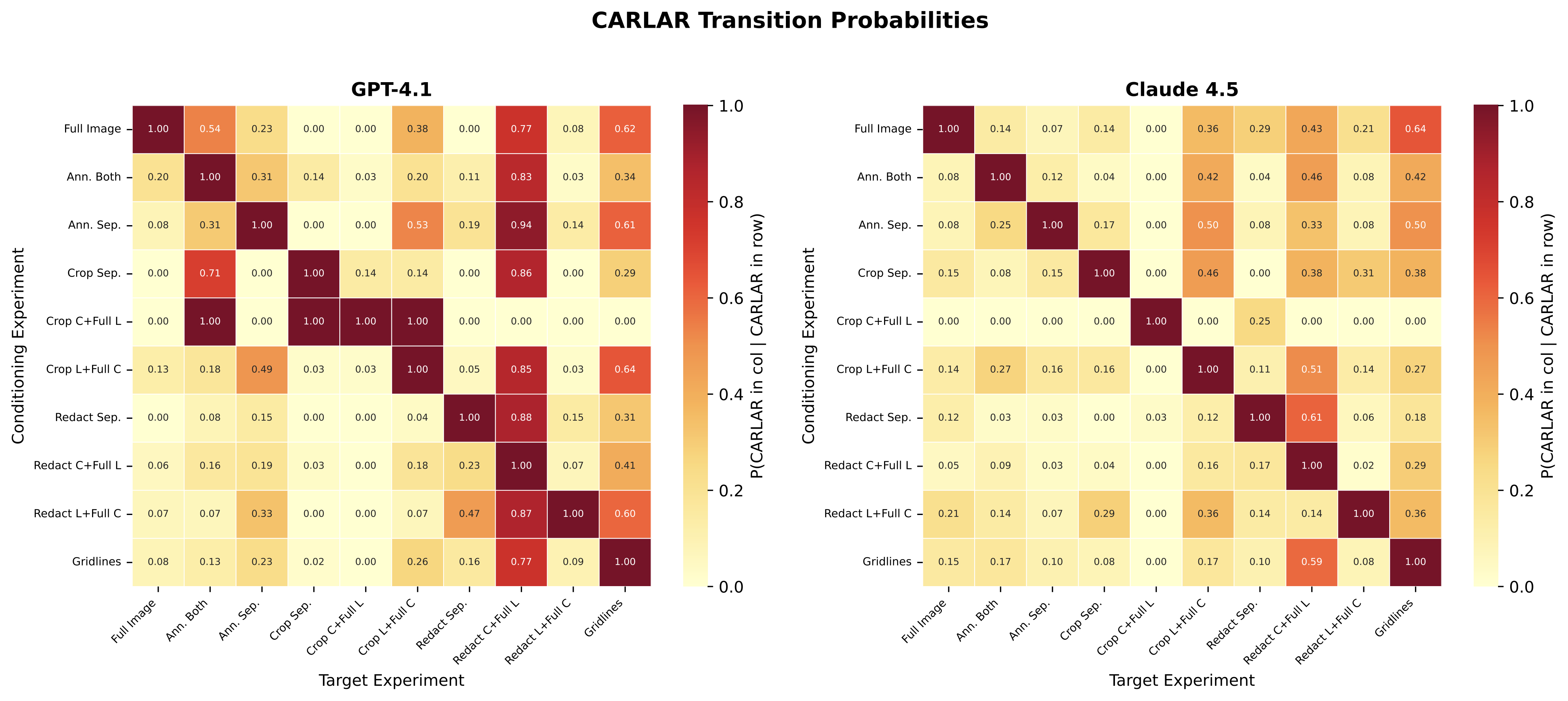}
    \caption{CARLAR transition probabilities: $P(\text{CARLAR in column} \mid \text{CARLAR in row})$. Higher values (darker cells) indicate stronger predictive relationships between experimental conditions.}
    \label{fig:carlar_heatmap}
\end{figure}

The transition heatmap (Figure~\ref{fig:carlar_heatmap}) shows asymmetric overlap between conditions. For GPT, 76.9\% of baseline CARLAR cases also exhibit CARLAR in Exp~8, but only 5.6\% of Exp~8 CARLAR cases do so at baseline; for Claude these conditional proportions are 42.9\% and 5.0\%. Redacting courtesy therefore elicits agreement in many cheques that do not exhibit it under full-image prompting.

\subsection{Compressed Image $n$-Shot for Correctness improvement and CARLAR mitigation}

\begin{table*}[t]
\centering
\small
\begin{tabular}{|l|c|c|c|c|c|c|}
\hline
\textbf{Experiment} & \textbf{Model} & \textbf{Court. Acc.} & \textbf{Legal Acc.} & \textbf{Both Acc.} & \textbf{CARLAR} & \textbf{CARLAR-NZ} \\
\hline
\textbf{Prompt-v1 Baseline} & GPT-4.1 & 85.2\% & 75.3\% & 64.6\% & 2.9\% & 2.9\% \\
\hline
Control Set & GPT-4.1 & 82.7\% & 88.9\% & 73.7\% & 9.5\% & 9.1\% \\
\hline
Prediction-Guided (Type 1) & GPT-4.1 & 79.8\% & 77.0\% & 63.4\% & 13.6\% & 11.9\% \\
\hline
Comprehensive Set & GPT-4.1 & 81.5\% & 85.2\% & 69.1\% & 9.1\% & 8.2\% \\
\hline
\textbf{Prompt-v1 Baseline} & Claude-4.5 & 54.7\% & 72.4\% & 41.2\% & 7.0\% & 4.1\% \\
\hline
Control Set & Claude-4.5 & 67.5\% & 75.7\% & 52.7\% & 0.4\% & 0.4\% \\
\hline
Prediction-Guided (Type 1) & Claude-4.5 & 63.8\% & 75.3\% & 48.1\% & 4.5\% & 0.8\% \\
\hline
Comprehensive Set & Claude-4.5 & 64.2\% & 75.7\% & 51.0\% & 2.1\% & 0.4\% \\
\hline
\end{tabular}
\caption{Prompt-v1 baseline and three verified compressed-image example sets: Prediction-Guided Type 1 ($n{=}7$), Control Type 2 ($n{=}10$), and Comprehensive Type 3 ($n{=}5$). All columns use the same 243 matched cheque IDs; nine non-cheque rows from the historical 252-row selection were excluded. These example sets differ from the confidence-elicitation prompts in Table~\ref{tab:mean_confidence}. Prompt-v1 and the full-image visual baseline use different prompts.}
\label{tab:fewshot_accuracy}
\end{table*}

Table~\ref{tab:fewshot_accuracy} presents run-matched results on 243 cheque IDs for prompt-v1 and all three verified example sets. Its detailed courtesy-first prompt differs from the visual-ablation full-image baseline in Table~\ref{tab:carlar_rates}: GPT CARLAR rates are 2.9\% and 5.3\%, respectively (Claude: 7.0\% and 5.8\%). These are different prompting runs on the same cheques, not interchangeable baselines. The Prediction-Guided Type-1 accuracy experiment is distinct from the Error-Similar confidence-prompt experiments in Table~\ref{tab:mean_confidence}.

For GPT-4.1, prompt-v1 yields 85.2\% courtesy, 75.3\% legal and 64.6\% joint accuracy. Prediction-Guided Type 1 yields 63.4\% joint accuracy, below baseline; Control improves legal accuracy to 88.9\% and joint accuracy to 73.7\%, while Comprehensive reaches 85.2\% and 69.1\%, respectively. All three increase CARLAR relative to the 2.9\% baseline (13.6\%, 9.5\%, and 9.1\%, respectively), so a joint-accuracy gain does not imply reduced predicted-field agreement.

For Claude 4.5, prompt-v1 yields 54.7\% courtesy and 41.2\% joint accuracy. Prediction-Guided Type 1 raises these to 63.8\% and 48.1\%; Control raises them to 67.5\% and 52.7\% (an 11.5-point joint gain); and Comprehensive reaches 64.2\% and 51.0\%. All three reduce CARLAR relative to the 7.0\% baseline (4.5\%, 0.4\%, and 2.1\%, respectively). Paired joint-accuracy tests below assess which apparent gains are distinguishable from sampling variation.

Across the three verified example sets, predicted-field agreement decreases for Claude and increases for GPT-4.1, while the joint-accuracy effect varies by set and model. Because the sets differ in both example count and content, neither factor is isolated; Type 1's prediction-guided context cannot be directly attributed to example count alone.

\textbf{Strict-subset sensitivity.} On the same 124 cheques with two observed unequal amounts, GPT-4.1 CARLAR is 5.6\% [1.6, 10.5] with prompt-v1 alone, 15.3\% [8.9, 21.8] with Prediction-Guided Type 1, 8.9\% [4.0, 14.5] with Control, and 9.7\% [4.8, 15.3] with Comprehensive. The corresponding Claude 4.5 rates are 6.5\% [2.4, 11.3], 4.8\% [1.6, 8.9], 0.8\% [0.0, 2.4], and 3.2\% [0.8, 6.5], respectively (95\% bootstrap intervals, 10{,}000 resamples). The model-specific direction persists on strict mismatches, but these descriptive rates do not establish significance for CARLAR changes or prove which field, if either, was copied.

\begin{table*}[t]
\centering
\small
\begin{tabular}{|l|l|c|c|}
\hline
\textbf{Example Set} & \textbf{Model} & \textbf{Joint Accuracy Gain (pp) [95\% CI]} & \textbf{BH-adjusted $p$} \\
\hline
Prediction-Guided (Type 1) & GPT-4.1 & $-1.2$ [$-7.0$, 4.5] & 0.780 \\
Control (Type 2) & GPT-4.1 & 9.1 [3.3, 14.8] & 0.0077 \\
Comprehensive (Type 3) & GPT-4.1 & 4.5 [$-0.8$, 9.9] & 0.173 \\
Prediction-Guided (Type 1) & Claude 4.5 & 7.0 [1.6, 12.3] & 0.0279 \\
Control (Type 2) & Claude 4.5 & 11.5 [6.2, 16.9] & 0.00054 \\
Comprehensive (Type 3) & Claude 4.5 & 9.9 [4.9, 14.8] & 0.00081 \\
\hline
\end{tabular}
\caption{Paired joint-extraction accuracy differences against the model's prompt-v1 baseline on the same 243 cheque IDs. Confidence intervals are percentile bootstraps of paired per-cheque differences (10{,}000 resamples); exact two-sided McNemar $p$-values are Benjamini--Hochberg-adjusted over these six few-shot comparisons. Intervals themselves are not multiplicity-adjusted.}
\label{tab:fewshot_paired}
\end{table*}

Table~\ref{tab:fewshot_paired} distinguishes the strongest accuracy gains from variants with uncertain or negative effects. Claude's Control set has the largest paired gain (11.5 percentage points [6.2, 16.9], adjusted $p = 0.00054$); GPT's Control set gains 9.1 points [3.3, 14.8] (adjusted $p = 0.0077$). GPT's Prediction-Guided and Comprehensive differences are not significant in this six-test family. These tests concern joint correctness, not the significance of changes in CARLAR agreement.

\begin{table*}[t]
\centering
\small
\begin{tabular}{|l|l|c|c|}
\hline
\textbf{Example Set} & \textbf{Model} & \textbf{Strict Joint Gain (pp) [95\% CI]} & \textbf{BH-adjusted $p$} \\
\hline
Prediction-Guided (Type 1) & GPT-4.1 & 5.6 [$-3.2$, 14.5] & 0.332 \\
Control (Type 2) & GPT-4.1 & 23.4 [14.5, 32.3] & $<0.00001$ \\
Comprehensive (Type 3) & GPT-4.1 & 14.5 [5.6, 23.4] & 0.0088 \\
Prediction-Guided (Type 1) & Claude 4.5 & 6.5 [$-0.8$, 13.7] & 0.201 \\
Control (Type 2) & Claude 4.5 & 8.1 [0.8, 15.3] & 0.105 \\
Comprehensive (Type 3) & Claude 4.5 & 4.0 [$-2.4$, 10.5] & 0.332 \\
\hline
\end{tabular}
\caption{Exploratory paired joint-accuracy sensitivity on the same 124 cheques with two nonblank unequal annotations. Percentile bootstrap intervals use 10{,}000 paired resamples; exact two-sided McNemar $p$-values are BH-adjusted over this separate six-test strict family. Intervals themselves are not multiplicity-adjusted and need not have the same decision boundary as the adjusted tests.}
\label{tab:fewshot_strict_accuracy}
\end{table*}

On the strict two-field subset, GPT's Control gain is 23.4 points [14.5, 32.3] (adjusted $p < 0.00001$). Claude's Control gain is 8.1 points [0.8, 15.3], but its adjusted $p = 0.105$ does not establish significance in this smaller cohort. The broader 243-case Claude gain therefore should not be presented as independently confirmed on strict mismatches; neither analysis by itself determines whether an unannotated amount is physically absent from the cheque.

\subsection{Confidence Score thresholding and Compressed Image $n$-Shot for Confidence Recalibration}

\begin{table*}[t]
\centering
\small
\caption{Mean reported confidence on 243 cheque rows and on CARLAR cases within each separate confidence run. Unreported or nonnumeric confidence is encoded as zero; CARLAR-case counts differ between runs, so these subset means are not paired comparisons. The confidence-table example sets are distinct from the accuracy-table exemplar runs.}
\label{tab:mean_confidence}
\textbf{(a) All Rows}\\[4pt]
\begin{tabular}{|l|c|c|c|c|c|}
\hline
\textbf{Experiment} & \textbf{Scale} & \textbf{Claude Courtesy} & \textbf{GPT Courtesy} & \textbf{Claude Legal} & \textbf{GPT Legal} \\
\hline
LLM as Judge Confidence        & 0--100 & 75.9 & 83.7 & 66.5 & 74.5 \\
Azure OCR Confidence           & 0--100 & 67.5 & 64.7 & 31.5 & 33.9 \\
Few-Shot: Error-Similar        & 0--100 & 75.4 & 83.9 & 85.5 & 73.9 \\
Few-Shot: Control (Non-Similar)& 0--100 & 71.9 & 84.6 & 84.0 & 72.1 \\
Prob.\ Few-Shot: Error-Similar & 0--1   & 0.7  & 0.6  & 0.8  & 0.6  \\
Prob.\ Few-Shot: Non-Similar   & 0--1   & 0.8  & 0.6  & 0.9  & 0.5  \\
\hline
\end{tabular}

\vspace{8pt}

\textbf{(b) CARLAR Cases Only}\\[4pt]
\begin{tabular}{|l|c|c|c|c|c|}
\hline
\textbf{Experiment} & \textbf{Scale} & \textbf{Claude Courtesy} & \textbf{GPT Courtesy} & \textbf{Claude Legal} & \textbf{GPT Legal} \\
\hline
LLM as Judge Confidence        & 0--100 & 65.2 & 94.9 & 79.0 & 94.3 \\
Azure OCR Confidence           & 0--100 & 4.8  & 0.0  & 7.0  & 0.0  \\
Few-Shot: Error-Similar        & 0--100 & 78.1 & 94.6 & 85.2 & 93.2 \\
Few-Shot: Control (Non-Similar)& 0--100 & 70.2 & 93.4 & 78.2 & 94.4 \\
Prob.\ Few-Shot: Error-Similar & 0--1   & 0.7  & 0.4  & 0.7  & 0.4  \\
Prob.\ Few-Shot: Non-Similar   & 0--1   & 0.8  & 0.4  & 0.8  & 0.4  \\
\hline
\end{tabular}
\end{table*}

We compare LLM-as-Judge, Azure OCR, and two confidence-oriented table prompts across the same 243 cheque images. These confidence-table prompts are separate experiments from the accuracy-table exemplars, and each CARLAR-only mean in Table~\ref{tab:mean_confidence}(b) conditions on that run's own predicted-field agreement; those subsets need not contain the same images.

GPT-4.1's LLM-as-Judge confidence averages 94.9 (courtesy) and 94.3 (legal) among its 55 CARLAR cases, versus 83.7 and 74.5 overall. Claude's corresponding courtesy mean is 65.2 among 44 CARLAR cases versus 75.9 overall. OCR reports near-zero mean confidence on its own CARLAR subsets, but this alone is not evidence of better calibration: a zero-confidence prediction can still be correct when a field is genuinely absent. Likewise, probability-scale means include zero or unreported responses; they do not support a claim that nearly every prediction falls in a high-confidence bin. The correctness-based calibration results below provide the appropriate comparison.

\subsection{Formal Calibration Analysis}

To move beyond descriptive confidence statistics, we evaluate model calibration using standard metrics: Expected Calibration Error (ECE), Brier score, and risk-coverage analysis. For each sample, we define a binary correctness indicator ($y_i = 1$ if prediction matches ground truth, $0$ otherwise) and pair it with the model's reported confidence $c_i \in [0,1]$.

\begin{table}[t]
\centering
\small
\caption{Calibration metrics for LLM-as-Judge and Azure OCR confidence on 243 cheques. Overall ECE uses 10 bins; CARLAR-only ECE uses five bins on separate run-specific subsets (GPT: 55 LLM, 22 OCR; Claude: 44 LLM, 40 OCR). Lower values indicate better calibration within each setting; ECE across different bin counts should not be compared as a paired change.}
\label{tab:calibration}
\begin{tabular}{|l|l|c|c|c|c|}
\hline
\textbf{Method} & \textbf{Field} & \multicolumn{2}{c|}{\textbf{ECE}} & \multicolumn{2}{c|}{\textbf{Brier}} \\
\cline{3-6}
& & \textbf{GPT} & \textbf{Claude} & \textbf{GPT} & \textbf{Claude} \\
\hline
\multicolumn{6}{|c|}{\textit{All Rows ($n = 243$)}} \\
\hline
LLM-as-Judge & Courtesy & 0.43 & 0.65 & 0.43 & 0.62 \\
LLM-as-Judge & Legal    & 0.28 & 0.27 & 0.31 & 0.34 \\
Azure OCR    & Courtesy & 0.31 & 0.30 & 0.32 & 0.33 \\
Azure OCR    & Legal    & 0.48 & 0.37 & 0.42 & 0.31 \\
\hline
\multicolumn{6}{|c|}{\textit{CARLAR Cases Only}} \\
\hline
LLM-as-Judge & Courtesy & \textbf{0.93} & 0.63 & \textbf{0.88} & 0.55 \\
LLM-as-Judge & Legal    & 0.05 & 0.07 & 0.10 & 0.11 \\
Azure OCR    & Courtesy & 0.82 & \textbf{0.97} & 0.82 & \textbf{0.97} \\
Azure OCR    & Legal    & 0.14 & 0.01 & 0.14 & 0.00 \\
\hline
\end{tabular}
\end{table}

Table~\ref{tab:calibration} presents ECE and Brier scores. For GPT-4.1 LLM-as-Judge courtesy confidence, ten-bin overall ECE is 0.43; five-bin ECE on its CARLAR subset is 0.93 (Brier score 0.88). Claude's corresponding ECE values are 0.65 overall and 0.63 on its CARLAR subset. The differing bin counts and subset sizes preclude interpreting these as a directly comparable before/after change. These CARLAR cases include cheques with blank raw annotations under the primary zero convention; the measures quantify calibration against the annotated values, not the causal source of an error.

Azure OCR has lower all-row courtesy ECE (0.31 GPT, 0.30 Claude), but its CARLAR-only courtesy ECE is high (0.82 and 0.97), and the OCR- and LLM-defined CARLAR subsets differ. Low OCR confidence on those cases cannot by itself be described as well calibrated or as a validated CARLAR detector. Wilcoxon signed-rank tests find different LLM and OCR confidence distributions ($p < 0.001$ for both fields and models); distributional difference alone does not establish which score detects cross-field substitution.

\begin{table}[t]
\centering
\scriptsize
\caption{Courtesy-field accuracy before and after filtering at the 0.9 confidence threshold. Overall accuracy, coverage, and selective accuracy are computed within each separate confidence experiment on the same 243 cheques; these are not the prompt-v1 predictions in Table~\ref{tab:fewshot_accuracy}.}
\label{tab:selective_accuracy}
\begin{tabular}{|l|l|c|c|c|}
\hline
\textbf{Method} & \textbf{Model} & \textbf{Overall Acc.} & \textbf{Coverage} & \textbf{Sel.\ Acc.} \\
\hline
LLM-as-Judge & GPT-4.1   & 67.9\% & 85.6\% & 63.5\% \\
LLM-as-Judge & Claude 4.5 & 38.3\% & 66.3\% & 35.4\% \\
Azure OCR    & GPT-4.1   & 92.2\% & 58.0\% & \textbf{98.6\%} \\
Azure OCR    & Claude 4.5 & 93.0\% & 60.9\% & \textbf{95.9\%} \\
\hline
\end{tabular}
\end{table}

\begin{figure}[t]
    \centering
    \includegraphics[width=\columnwidth]{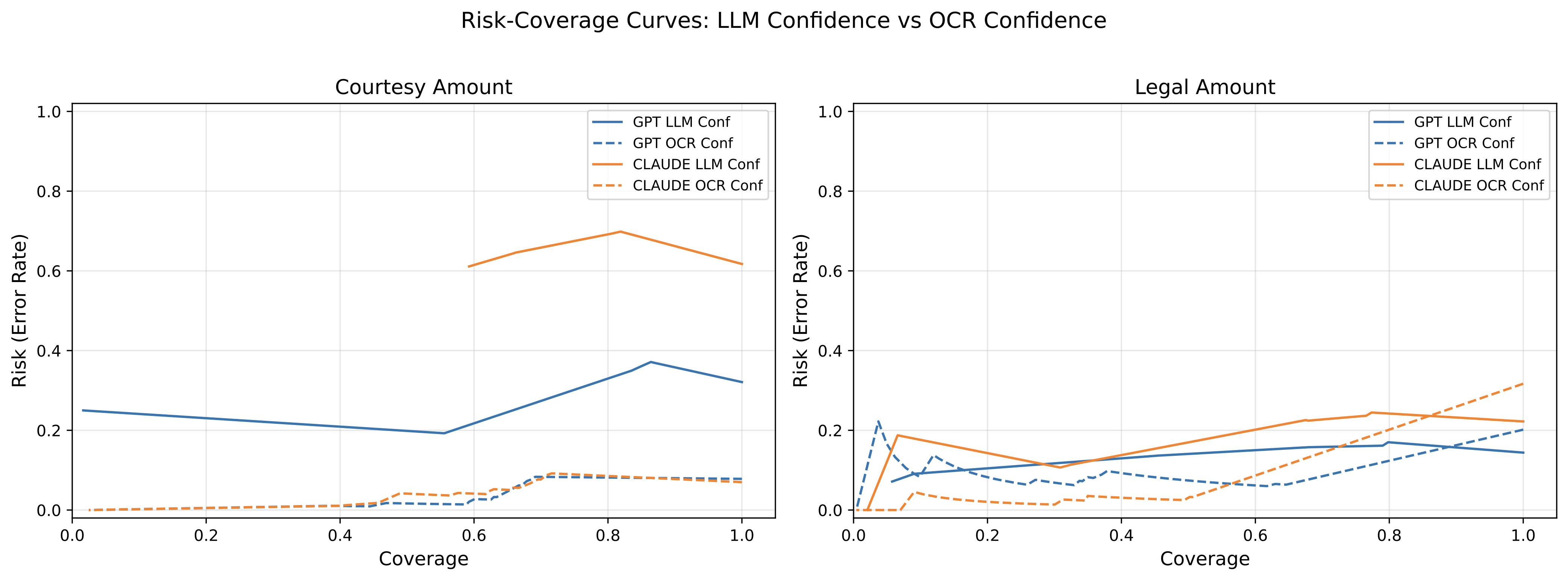}
    \caption{Risk-coverage curves for courtesy (left) and legal (right) amount extraction. LLM-as-Judge confidence (solid lines) yields nearly flat risk curves, while OCR confidence (dashed) shows a clear risk--coverage tradeoff.}
    \label{fig:risk_coverage}
\end{figure}

Table~\ref{tab:selective_accuracy} and Figure~\ref{fig:risk_coverage} quantify whether confidence thresholding can filter unreliable predictions. The unfiltered accuracies come from the 243-cheque confidence-elicitation runs, not the prompt-v1 accuracy baseline. At the 0.9 threshold, GPT-4.1's LLM-as-Judge confidence retains 85.6\% of cases but reduces courtesy accuracy from 67.9\% overall to 63.5\% among accepted cases (Claude: 38.3\% to 35.4\% at 66.3\% coverage). Azure OCR retains 58.0\% of GPT cases and raises accuracy from 92.2\% overall to 98.6\% (Claude: 93.0\% to 95.9\% at 60.9\% coverage). These are within-run comparisons, not a head-to-head extraction comparison under an identical prompt. General selective accuracy does not measure CARLAR-specific recall or prove that OCR is a reliable escalation trigger for cross-field substitution.

\section{Conclusion}

This work shows that multimodal LLMs can return agreeing amount predictions under visual ambiguity even when both annotations disagree or one raw annotation is blank. We call this measurable failure signature CARLAR influence; its possible mechanisms, including cross-field inference, warrant further testing before a causal interpretation. Our experiments reveal three principal findings.

First, unexpected predicted-field agreement occurs across three cheque datasets: among 722 strict two-field mismatches, full-image baseline rates range from 4.0\% to 9.7\%; the primary 243-cheque cohort also includes 119 blank-annotation cases. Redacting courtesy increases agreement sharply even on the strict subset (71.0\% GPT, 52.4\% Claude). RDA shows that non-CARLAR predictions can exhibit partial numeric drift, while 73.3\% of GPT cheques and 68.7\% of Claude cheques exhibit CARLAR in only one to three of the ten visual conditions. These findings concern sensitivity to input presentation, not a demonstrated copying mechanism in every case.

Second, the three verified compressed-image variants do not help uniformly. Claude's Control condition gives the largest paired gain on all 243 cheques, 11.5 percentage points [6.2, 16.9] (six-test BH-adjusted $p = 0.00054$); its strict-124 gain is 8.1 points with adjusted $p = 0.105$. GPT's Prediction-Guided Type-1 variant does not improve joint accuracy on the full cohort, whereas GPT's strict Control gain is 23.4 points [14.5, 32.3] (adjusted $p < 0.00001$). Predicted-field agreement falls for Claude but rises for GPT-4.1 across all three variants. We do not claim a universal CARLAR countermeasure.

Third, LLM-as-Judge confidence does not reliably filter courtesy errors at the tested threshold, whereas OCR confidence improves selective courtesy accuracy within its own extraction run. Whether either method detects CARLAR cases specifically remains untested; purpose-built calibration and detection for multi-field extraction remain open challenges.

Overall, reliable deployment of multimodal LLMs in high-stakes domains requires context-aware prompting, principled input design, and validation mechanisms that account for cross-field dependencies and model-specific biases.

\section{Future Directions}

Several promising directions emerge from this work. Our baseline cross-dataset evaluation observes predicted-field agreement in three cheque corpora spanning U.S., Singaporean, and Indian formats (722 strict mismatches in total), supporting limited within-domain generalizability. CARLAR-like influence likely extends further to other structured extraction tasks with correlated fields --- invoices, forms, or medical records --- and to a broader range of document formats and models. Wider geographic coverage could reveal whether CARLAR's persistence and the remedy's effectiveness are format-dependent, given that models may carry inherent biases toward the formats they were trained on. Additionally, access to open-source models was constrained by our institution's existing LLM infrastructure; future work extending this analysis to open-source models would enable direct fine-tuning and weight-level investigation of the underlying mechanism.

We acknowledge several limitations in the current example-selection design. As noted in Section~4.2, the single-image constraint couples $n$ with content type (the breadth-vs-depth tradeoff), so their effects are not independently isolated. Example selection was not randomized or repeated: a single fixed set of cheques was used per condition. Example and evaluation image IDs were compared and had no overlap, even though both were drawn from the broader public train split. Randomized example selection from a separately held-out pool with repeated draws would provide stronger evidence about generality. All three example-set conditions have matched-ID prediction outputs, but they differ in content as well as number; observed model-specific CARLAR effects therefore cannot be attributed to either factor in isolation.

A further direction is a systematic ablation over the number of examples $n$ in the compressed image few-shot remedy, disentangling it from content variation. Too few examples may under-specify the desired reasoning pattern, while too many may exceed the model's effective context capacity or introduce noise; given that GPT-4.1 and Claude 4.5 already respond differently to example selection strategy, their sensitivity to $n$ likely also differs. Future work should ablate $n$ across prediction-guided, control, and comprehensive example sets while holding content type fixed, to measure its isolated effect on joint accuracy, CARLAR, and CARLAR-NZ, and to determine whether performance gains plateau quickly or scale with richer example sets. Such analysis would inform principled example-count selection for deploying this remedy reliably in production settings.

Finally, improving confidence calibration and example selection for structured extraction tasks remain open challenges. Existing calibration techniques are largely designed for classification settings and may not transfer effectively to multi-field generation with inter-field dependencies, while the verified prediction-guided, control, and comprehensive variants have different model-specific effects — motivating retrieval strategies that balance diversity, relevance, and stability. Jointly addressing calibration and example selection could meaningfully improve both accuracy and deployment safety in real-world financial workflows.

\bibliographystyle{ACM-Reference-Format}
\bibliography{main}

@misc{openai_chatgpt4.1,
  title        = {GPT-4.1},
  author       = {OpenAI},
  howpublished = {\url{https://platform.openai.com/docs/models/gpt-4.1}},
  year         = {2025},
  note         = {Accessed: 2026-01-20}
}

@misc{claude_sonnet_4.5,
  title        = {Claude-sonnet-4.5},
  author       = {Anthropic},
  howpublished = {\url{https://www.anthropic.com/news/claude-4-5-sonnet}},
  year         = {2025},
  note         = {Accessed: 2026-01-20}
}

@inproceedings{mathew2021docvqa,
  title        = {{DocVQA}: A Dataset for {VQA} on Document Images},
  author       = {Mathew, Minesh and Karatzas, Dimosthenis and Jawahar, C.V.},
  booktitle    = {Proceedings of the IEEE/CVF Winter Conference on Applications of Computer Vision (WACV)},
  pages        = {2200--2209},
  year         = {2021}
}

@article{zhou2022coop,
  title        = {Learning to Prompt for Vision-Language Models},
  author       = {Zhou, Kaiyang and Yang, Jingkang and Loy, Chen Change and Liu, Ziwei},
  journal      = {International Journal of Computer Vision},
  volume       = {130},
  number       = {9},
  pages        = {2337--2348},
  year         = {2022},
  doi          = {10.1007/s11263-022-01653-1}
}

@inproceedings{brown2020gpt3,
  title        = {Language Models are Few-Shot Learners},
  author       = {Brown, Tom and Mann, Benjamin and Ryder, Nick and Subbiah, Melanie and Kaplan, Jared and Dhariwal, Prafulla and Neelakantan, Arvind and Shyam, Pranav and Sastry, Girish and Askell, Amanda and others},
  booktitle    = {Advances in Neural Information Processing Systems},
  volume       = {33},
  pages        = {1877--1901},
  year         = {2020}
}

@article{raffel2020t5,
  title        = {Exploring the Limits of Transfer Learning with a Unified Text-to-Text Transformer},
  author       = {Raffel, Colin and Shazeer, Noam and Roberts, Adam and Lee, Katherine and Narang, Sharan and Matena, Michael and Zhou, Yanqi and Li, Wei and Liu, Peter J.},
  journal      = {Journal of Machine Learning Research},
  volume       = {21},
  number       = {140},
  pages        = {1--67},
  year         = {2020}
}

@article{huang2025hallucination,
  title        = {A Survey on Hallucination in Large Language Models: Principles, Taxonomy, Challenges, and Open Questions},
  author       = {Huang, Lei and Yu, Weijiang and Ma, Weitao and Zhong, Weihong and Feng, Zhangyin and Wang, Haotian and Chen, Qianglong and Peng, Weihua and Feng, Xiaocheng and Qin, Bing and Liu, Ting},
  journal      = {ACM Transactions on Information Systems},
  volume       = {43},
  number       = {2},
  articleno    = {42},
  pages        = {1--55},
  year         = {2025},
  doi          = {10.1145/3703155}
}

@inproceedings{li2023visionhallucination,
  title        = {Evaluating Object Hallucination in Large Vision-Language Models},
  author       = {Li, Yifan and Du, Yifan and Zhou, Kun and Wang, Jinpeng and Zhao, Wayne Xin and Wen, Ji-Rong},
  booktitle    = {Proceedings of the 2023 Conference on Empirical Methods in Natural Language Processing (EMNLP)},
  pages        = {292--305},
  year         = {2023},
  publisher    = {Association for Computational Linguistics},
  doi          = {10.18653/v1/2023.emnlp-main.20}
}

@inproceedings{minderer2021confidence,
  title        = {Revisiting the Calibration of Modern Neural Networks},
  author       = {Minderer, Matthias and Djolonga, Josip and Romijnders, Rob and Hubis, Frances and Zhai, Xiaohua and Houlsby, Neil and Tran, Dustin and Lucic, Mario},
  booktitle    = {Advances in Neural Information Processing Systems},
  volume       = {34},
  year         = {2021}
}

@inproceedings{geifman2017selective,
  title        = {Selective Classification for Deep Neural Networks},
  author       = {Geifman, Yonatan and El-Yaniv, Ran},
  booktitle    = {Advances in Neural Information Processing Systems},
  volume       = {30},
  year         = {2017}
}

@misc{verma2024handwrittenchequevqa,
  title        = {Handwritten Cheque VQA Dataset},
  author       = {Verma, Aniket},
  howpublished = {\url{https://huggingface.co/datasets/aniketVerma07/handwritten_cheque_vqa_dataset}},
  year         = {2024},
  note         = {Hugging Face dataset, Accessed: 2026-03-20}
}

@misc{arunchincheit2024cheque,
  title        = {Cheque Dataset},
  author       = {Arunchincheit},
  howpublished = {\url{https://huggingface.co/datasets/arunchincheit/cheque}},
  year         = {2024},
  note         = {Hugging Face dataset, Accessed: 2026-10-04}
}

@misc{shivalikasingh2024cheque,
  title        = {Cheque Dataset},
  author       = {Singh, Shivalika},
  howpublished = {\url{https://huggingface.co/datasets/shivalikasingh/cheque}},
  year         = {2024},
  note         = {Hugging Face dataset, Accessed: 2026-10-04}
}

@inproceedings{groot2024overconfidence,
  title        = {Overconfidence is Key: Verbalized Uncertainty Evaluation in Large Language and Vision-Language Models},
  author       = {Groot, Tobias and Valdenegro-Toro, Matias},
  booktitle    = {Proceedings of the 4th Workshop on Trustworthy Natural Language Processing (TrustNLP)},
  pages        = {145--171},
  year         = {2024},
  publisher    = {Association for Computational Linguistics},
  doi          = {10.18653/v1/2024.trustnlp-1.13}
}

@misc{kadavath2022confidence,
  title        = {Language Models (Mostly) Know What They Know},
  author       = {Kadavath, Saurav and Conerly, Tom and Askell, Amanda and Henighan, Tom and Drain, Dawn and Perez, Ethan and Schiefer, Nicholas and Hatfield-Dodds, Zac and DasSarma, Nova and Tran-Johnson, Eli and others},
  year         = {2022},
  eprint       = {2207.05221},
  archivePrefix= {arXiv},
  primaryClass = {cs.CL},
  note         = {arXiv preprint}
}

@inproceedings{lin2022truthfulqa,
  title        = {{TruthfulQA}: Measuring How Models Mimic Human Falsehoods},
  author       = {Lin, Stephanie and Hilton, Jacob and Evans, Owain},
  booktitle    = {Proceedings of the 60th Annual Meeting of the Association for Computational Linguistics (Volume 1: Long Papers)},
  pages        = {3214--3252},
  year         = {2022},
  publisher    = {Association for Computational Linguistics},
  doi          = {10.18653/v1/2022.acl-long.229}
}

@inproceedings{xu2020layoutlm,
  title        = {{LayoutLM}: Pre-training of Text and Layout for Document Image Understanding},
  author       = {Xu, Yiheng and Li, Minghao and Cui, Lei and Huang, Shaohan and Wei, Furu and Zhou, Ming},
  booktitle    = {Proceedings of the 26th ACM SIGKDD International Conference on Knowledge Discovery and Data Mining},
  pages        = {1192--1200},
  year         = {2020},
  publisher    = {ACM}
}

@inproceedings{hu2023promptcap,
  title        = {{PromptCap}: Prompt-Guided Image Captioning for {VQA} with {GPT-3}},
  author       = {Hu, Yushi and Hua, Hang and Yang, Zhengyuan and Shi, Weijia and Smith, Noah A. and Luo, Jiebo},
  booktitle    = {Proceedings of the IEEE/CVF International Conference on Computer Vision (ICCV)},
  pages        = {2963--2975},
  year         = {2023}
}

@article{awal_vqa_prompting_2023,
  title        = {Investigating Prompting Techniques for Zero- and Few-Shot Visual Question Answering},
  author       = {Awal, Rabiul and Zhang, Le and Agrawal, Aishwarya},
  journal      = {arXiv preprint arXiv:2306.09996},
  year         = {2023}
}

@misc{kong2026evaluating,
  title        = {Evaluating LLMs in Finance Requires Explicit Bias Consideration},
  author       = {Kong, Yaxuan and Lee, Hoyoung and Hwang, Yoontae and Lopez-Lira, Alejandro and Levy, Bradford and Mehta, Dhagash and Wen, Qingsong and Choi, Chanyeol and Lee, Yongjae and Zohren, Stefan},
  year         = {2026},
  eprint       = {2602.14233},
  archivePrefix= {arXiv},
  primaryClass = {cs.LG},
  note         = {arXiv preprint}
}

@inproceedings{thakur2025judging,
  title        = {Judging the Judges: Evaluating Alignment and Vulnerabilities in {LLMs}-as-Judges},
  author       = {Thakur, Aman Singh and Choudhary, Kartik and Ramayapally, Venkat Srinik and Vaidyanathan, Sankaran and Hupkes, Dieuwke},
  booktitle    = {Proceedings of the Fourth Workshop on Generation, Evaluation and Metrics (GEM\textsuperscript{2})},
  pages        = {404--430},
  year         = {2025},
  publisher    = {Association for Computational Linguistics}
}

@inproceedings{naeini2015ece,
  title        = {Obtaining Well Calibrated Probabilities Using {B}ayesian Binning},
  author       = {Naeini, Mahdi Pakdaman and Cooper, Gregory F. and Hauskrecht, Milos},
  booktitle    = {Proceedings of the AAAI Conference on Artificial Intelligence},
  volume       = {29},
  pages        = {2901--2907},
  year         = {2015},
  publisher    = {AAAI Press}
}

@article{brier1950verification,
  title        = {Verification of Forecasts Expressed in Terms of Probability},
  author       = {Brier, Glenn W.},
  journal      = {Monthly Weather Review},
  volume       = {78},
  number       = {1},
  pages        = {1--3},
  year         = {1950}
}

@inproceedings{guo2017calibration,
  title        = {On Calibration of Modern Neural Networks},
  author       = {Guo, Chuan and Pleiss, Geoff and Sun, Yu and Weinberger, Kilian Q.},
  booktitle    = {Proceedings of the 34th International Conference on Machine Learning (ICML)},
  pages        = {1321--1330},
  year         = {2017}
}

\clearpage
\section*{Appendix}

\appendix

\begin{figure}[H]
    \centering
    \includegraphics[width=0.65\linewidth]{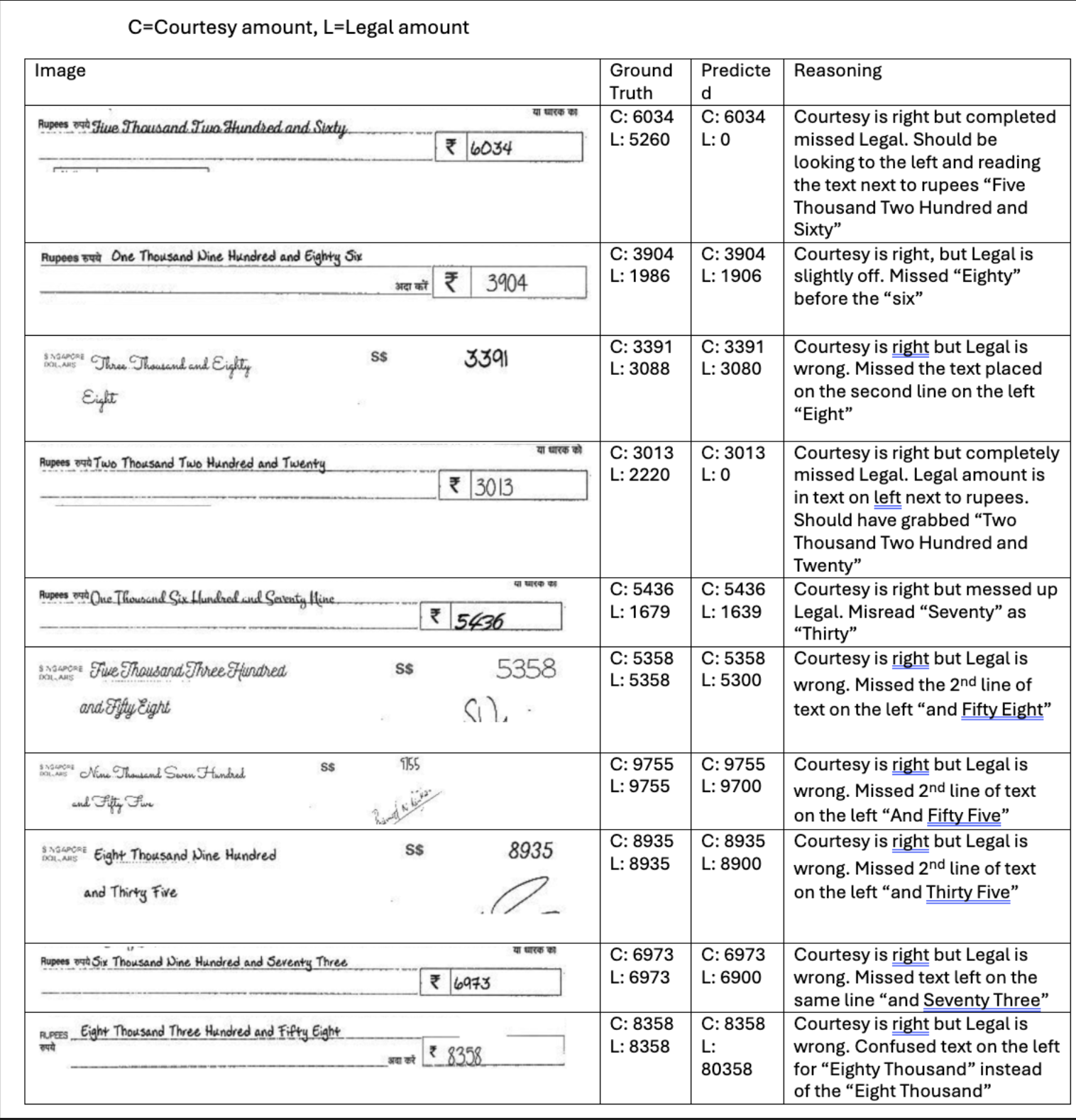}
    \caption{Compressed screenshot of Few-shot table}
    \label{fig:nshot_Table}
\end{figure}

\end{document}